\documentclass{article}

\PassOptionsToPackage{numbers, compress}{natbib}

\usepackage[preprint]{neurips_2024}

\usepackage{amsmath}
\usepackage{algorithm}
\usepackage{algpseudocode}
\usepackage{graphicx}
\usepackage{booktabs}
\usepackage{tabularx}
\usepackage{amsmath}
\usepackage{amssymb}
\usepackage{afterpage}
\usepackage[colorinlistoftodos]{todonotes}
\usepackage{wrapfig,lipsum,booktabs}

\usepackage[utf8]{inputenc} 
\usepackage[T1]{fontenc}    
\usepackage{hyperref}       
\usepackage{url}            
\usepackage{booktabs}       
\usepackage{amsfonts}       
\usepackage{nicefrac}       
\usepackage{microtype}      
\usepackage{xcolor}         
\usepackage{amsmath}
\usepackage{color, colortbl}
\usepackage{fontawesome5}

\title{Image-of-Thought Prompting for Visual Reasoning Refinement in Multimodal Large Language Models}

\author{%
Qiji Zhou$^1\thanks{Equal Contribution.}$ \quad Ruochen Zhou$^{3*}$ \quad Zike Hu$^{4*}$ \quad Panzhong Lu$^1$ \quad \\
\textbf{Siyang Gao}$^3$ \quad \textbf{Yue Zhang}$^{2,1\thanks{Corresponding Author.}}$\\
$^1$School of Engineering, Westlake University\\ $^2$Research Center for Industries of the Future, Westlake University\\ $^3$City University of Hong Kong \quad $^4$ Johns Hopkins University
}

\begin{document}

\maketitle

\begin{abstract}
Recent advancements in Chain-of-Thought (CoT) and related rationale-based works have significantly improved the performance of  Large Language Models (LLMs)  in complex reasoning tasks.
With the evolution of Multimodal Large Language Models (MLLMs), enhancing their capability to tackle complex multimodal reasoning problems is a crucial frontier.
However, incorporating multimodal rationales in CoT has yet to be thoroughly investigated.
We propose the Image-of-Thought (IoT) prompting method, which helps MLLMs to extract visual rationales step-by-step.
Specifically, IoT prompting can automatically design critical visual information extraction operations based on the input images and questions.
Each step of visual information refinement identifies specific visual rationales that support answers to complex visual reasoning questions.
Beyond the textual CoT, IoT simultaneously utilizes visual and textual rationales to help MLLMs understand complex multimodal information.
IoT prompting has improved zero-shot visual reasoning performance across various visual understanding tasks in different MLLMs. Moreover, the step-by-step visual feature explanations generated by IoT prompting elucidate the visual reasoning process, aiding in analyzing the cognitive processes of large multimodal models.
\end{abstract}

\section{Introduction}
\begin{figure}[h]
    \centering
    \includegraphics[width=0.9\linewidth]{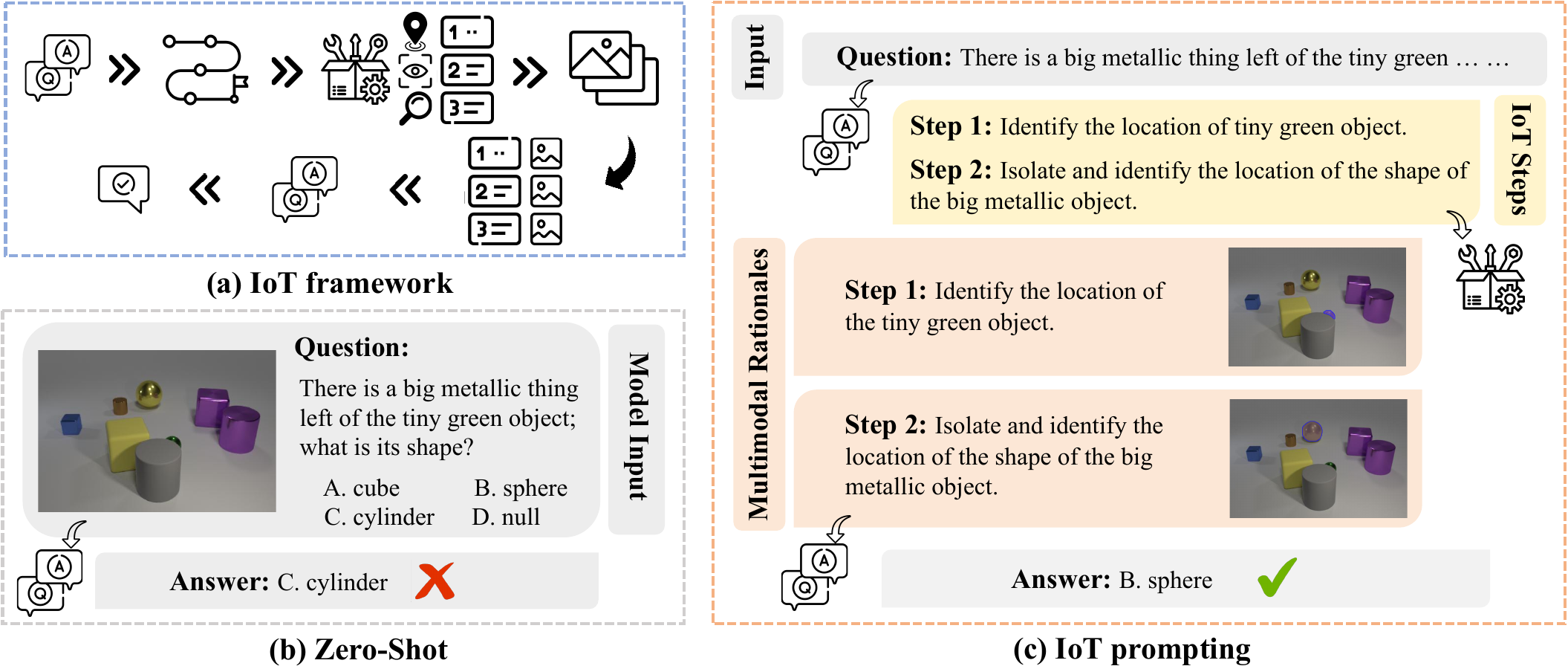}
    \caption{(a) illustrates the overall framework of our method. Zero-shot and our IoT method are compared in (b) and (c).}
    \label{fig:fig_sys}
\end{figure}

The emergence of Large Language Models (LLMs)\citep{brown2020language,ouyang2022training,touvron2023llama,chung2024scaling,chowdhery2023palm}~equipped with train-free\citep{wei2022chain,kojima2022large,zhang2022automatic,huang2024free,tang2023cotdet}~or fine-tuning\citep{magister2022teaching,ho2022large,li2022advance,wang2022self}~by the Chain-of-Thought (CoT) prompting techniques has marked a significant advancement in the field of complex reasoning. 
This technique not only enhances model performance but also improves interpretability by making the inner reasoning process transparent through textual rationales\citep{zhao2024explainability,wei2022chain,zheng2023ddcot,jie2024interpretable}. As shown in Figure \ref{fig:fig_sys} (b), without CoT, the model might make mistakes in questions with complex requirements. However, when model is asked to split the reasoning process into different steps, it guides the model to follow a logical chain, such as identifying and locating the aimed objects in the question, thus arriving at the correct answer.

In general, using language to memorize and reason is a natural human habit. However, relying solely on textual rationale 
to reason about multimodal data is neither perceptual nor direct. Specifically, when we tackle figure or geometry questions, we often intuitively construct our chain-of-thought on multimodal data (include text and images). To facilitate our thinking, we often create drafts or mark figures to deal with each step directly rather than only transferring visual information to text and think through the text. Thus, inspired by that, we develop a new method that mimics the human mind, as illustrated in Figure \ref{fig:fig_sys} (c). 

Some current works already tried to bridge the gap between modalities and reduce the errors from LLMs' overreliance on linguistic knowledge, especially within the Large Vision-Language Models (VLMs)\citep{liu2023improved,achiam2023gpt,geminiteam2024gemini}. 
MMCoT\citep{zhang2023multimodal} necessitates training to generate intermediate reasoning steps, which may serve as explanations accompanied by answers or as reasoning prior to inferring an answer. 
Building on this foundation, DDCoT\citep{zheng2023ddcot} decomposes each question into subquestions and employs external Visual Question-Answering (VQA) models to address these subcomponents. However, these methods primarily generate textual rationales and suffer from superficial integration, where visual data is not fully synthesized with textual reasoning. And this often leads to hallucination, where the models generate plausible but incorrect or unsupported content. 
Such errors highlight the need for a more integrated approach that combines these modalities to reflect the complex nature of human cognitive processing. In addition, some works try to invoke external detection tools to translate multimodal information into single modality rationales\citep{tang2023cotdet,jiang2024joint}~or design an agent to call multiple tools following fixed template rules\citep{shaham2024multimodal,wu2024dettoolchain}. Nevertheless, these approaches fail to actualize the genuine essence of achieving automatic step-by-step thinking in MLLMs. Instead, they artificially dictate the necessary tools and predetermined sequence to be employed. On this basis, our method employs a train-free paradigm, which not only maintains the integrity of multimodal data but also utilizes the MLLM's ability to automatically design the extracting processes for the textual and visual rationales itself to enhance its ability of multimodal reasoning.

In this paper, we proposed the Image-of-Thought (IoT) prompting method, which explores how the MLLMs can leverage visual and textual rationales for complex visual reasoning tasks.
In other words, our method compels MLLMs to engage directly with images through a step-by-step reasoning process, anchoring decisions more firmly in visual reality rather than predominantly on textual interpretations. Initially, we enable the MLLM to design the visual and textual steps, guiding our IoT model to utilize external image processing tools to generate a multimodal rationale series. This rationale series subsequently assists the MLLM in deriving the answer, as illustrated in Figure~\ref{fig:fig_sys}.
By generating a multimodal rationale series that closely aligns with each step of the CoT-like process, our approach ensures that every reasoning step anchored the textual and visual rationale. This integration reduces the likelihood of hallucinations and diminishes the reliance on textual biases. Notably, this IoT method is designed with an MLLM-centric reasoning process, allowing the MLLM to guide the entire reasoning chain. This systematic integration of discrete image-processing operations within the model’s reasoning chains creates a more consistent and accurate multimodal reasoning framework. It enhances stability across reasoning tasks by employing a unified model to orchestrate the entire reasoning process, including step planning and the selection of image-processing tools. Based on our experiments, IoT prompting promises to mitigate issues commonly associated with hallucinations and boosts the model’s capacity to undertake complex reasoning tasks that mirror human cognitive processes.

Empirical evaluation of the IoT prompting across diverse benchmarks on the Visual Question-Answering task demonstrates its effectiveness in reducing errors associated with traditional multimodal CoT approaches. Notably, we are the first to work on generating multimodal rationales. Our method sets a new paradigm in multimodal reasoning that can be extended to any reasoning task with the use of extendable external tools. The training-free nature of the IoT paradigm eliminates the need for expensive and time-consuming fine-tuning processes typically required by other models, thereby enhancing both the accuracy and interpretability of MLLMs' reasoning processes. The multimodal rationale not only offers a well-rounded understanding of the internal reasoning process of MLLMs but also makes it more comprehensible to human observers.

\section{Related Work}

\textbf{CoT Reasoning with LLMs. }
Advances in natural language processing have demonstrated the effectiveness of Large Language Models (LLMs) using Chain-of-Thought (CoT) reasoning for enhanced problem-solving. CoT prompts LLMs to articulate intermediate reasoning steps, significantly boosting reasoning abilities. \citep{wei2022chain} and \citep{kojima2022large} have shown how simple prompts or a few detailed examples can significantly boost LLMs' reasoning in zero-shot and few-shot scenarios, respectively. Research has focused on optimizing these techniques by refining example selection based on similarity, diversity, and complexity \citep{rubin2021learning,lu2022dynamic,zhang2022automatic,fu2022complexity}, and incorporating structured methods like programming \citep{chen2022program}, problem decomposition \citep{khot2022decomposed,zhou2022least}, and rationale calibration \citep{wang2022self,li2022advance,fu2022complexity}.

\textbf{CoT Reasoning with MLLMs. }
Numerous investigations have verified the significant impact of Chain-of-thought prompting on enhancing of MLLMs. 
Zhang et al.\cite{zhang2023multimodal} leveraged visual information to produce relevant rationale that enhances the model's reasoning capability. Furthermore, Zheng et al.\cite{zheng2023ddcot} divided questions into sub-questions and utilized responses from the VQA model to generate rationales. Additionally, Zhang et al.\cite{zhang2024cocot} guided the model in addressing intricate queries concerning multiple image inputs by comparing the similarities and dissimilarities among the images with multiple inputs. Mitra et al.\cite{mitra2023compositional} pioneered the utilization of LLM in generating Scene Graphs and subsequently employed the model to generate answers.

\textbf{Visual Prompting in MLLMs. }
In addition to focusing on utilizing LLM for generating intermediate multi-step text inference processes, several studies have also explored the incorporation of visual features to introduce intermediate steps. Various techniques employing visual cues such as boxes and masks have been devised to enhance the model's performance in image inference tasks. Jiang et al.\cite{jiang2024joint} add an image containing a bounding box around the target object within the inference process to aid MLLMs in reasoning. They evaluated the zero-shot outcomes of this approach on GPT4 and Gemine-Pro. Additionally, Lin et al.\cite{lin2024draw} trained an MLLM to recognize image inputs with bounding boxes, thereby bolstering the model's reasoning capability. Likewise, Shao et al.\cite{shao2024visual} trained a model with an LLaVA structure, allowing simultaneous input of the original image and the image with bounding boxes, employing visual cues to enhance the model's reasoning ability. 

Based on these methods, we proposed IoT prompting, which can automatically generate the paired textual and visual rationales simultaneously.
Our work extends CoT reasoning to multimodal contexts, addressing inherent challenges and creating multimodal rationales, thus fostering a more holistic and accurate multimodal reasoning framework. 

\begin{figure}[h]
    \centering
    \includegraphics[width=0.9\textwidth]{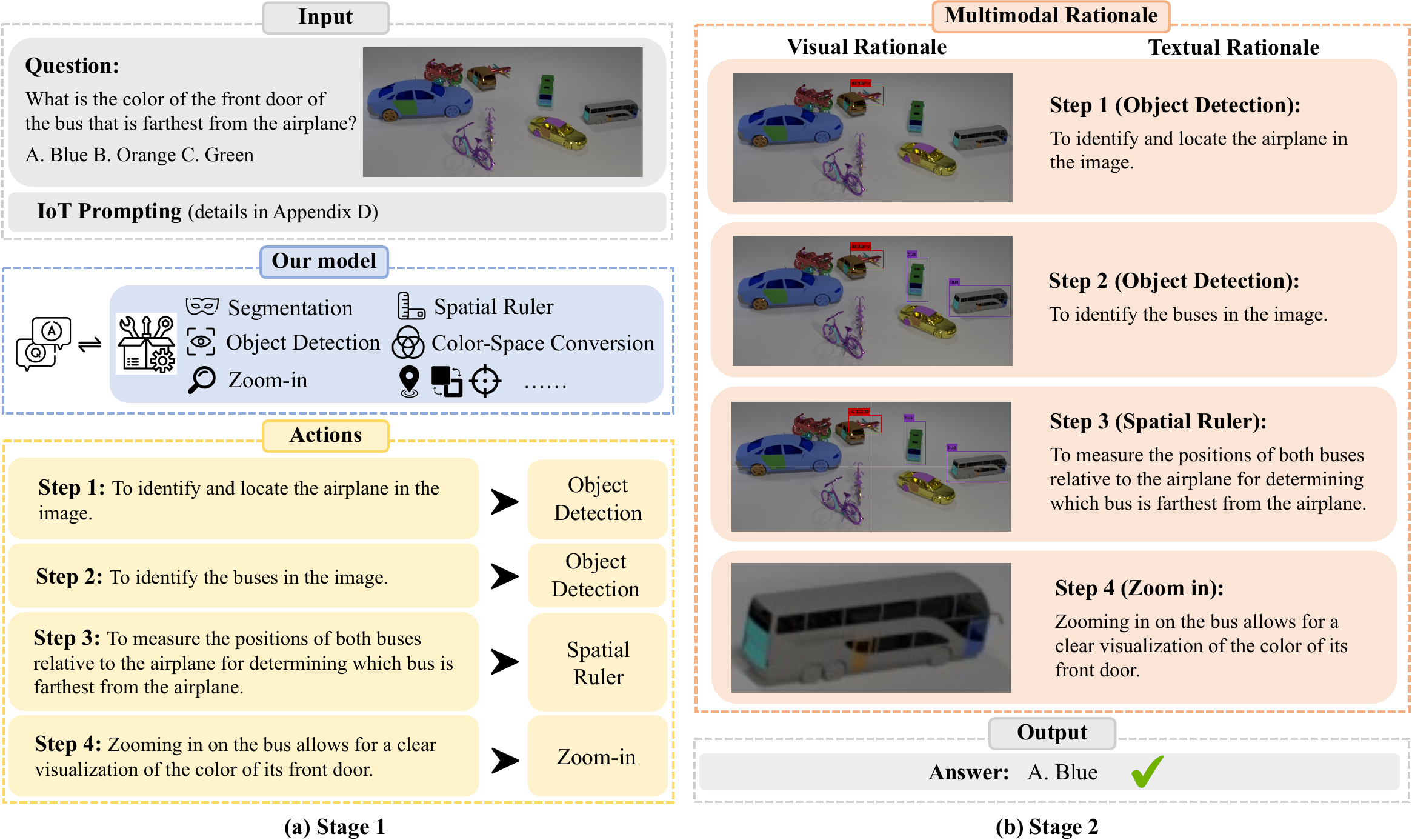}
    \caption{Overview of the IoT Method: (a) Stage 1 \textit{Chain-of-Multimodal-Rationales} and (b) Stage 2 \textit{Hybrid-Rationale-based Answer Refining}. }
    \label{fig:tools}
\end{figure}

\section{Method}
\subsection{Concept Definition}
\textbf{Multimodal Reasoning:}
This kind of task involves the integration of inputs from multiple modalities, such as images alongside textual information. In these tasks, the model is required to produce accurate responses $(A)$ to given questions $(Q)$ based on the provided multimodal inputs $(I)$. As illustrated in Figure~\ref{fig:fig_sys}, while the task involves comprehending complex spatial relationships within images, the model should accurately understand the shapes and positions of various objects to answer questions correctly.

\textbf{Hybrid Rationales:}
In multimodal reasoning tasks, hybrid rationales play a critical role in enhancing the interpretability and accuracy of the model's responses by combining textual and visual elements. The textual rationale $(TR)$ consists of step-by-step reasoning generated by MLLMs, which outlines the logical sequence the model follows to support its answers. Concurrently, the visual rationale $(VR)$ involves extracting relevant visual features from the original images, guided by $TR$. This ensures that the visual information directly corresponds to the logical steps outlined in the $TR$, creating a coherent and integrated explanation. By combining these textual and visual rationales, the model offers more robust and transparent justifications for its answers, as illustrated in the right part of Figure~\ref{fig:fig_sys}.

\subsection{Image-of-Thought Prompting}
In this chapter, we introduce the Image-of-Thought (IoT) prompting method.
Firstly, we show the chain of hybrid rationale generation.
Then, we illustrate the multimodal answer refining with hybrid rationales.

\subsubsection{Chain-of-Multimodal-Rationales}
\label{cpt2-1}
\textbf{MLLM-Centric Reasoning with Actions. }
To facilitate detailed visual reasoning, our method employs a structured approach where the MLLM autonomously plans and executes a sequence of image processing actions $(Action)$. The MLLM divides problems into distinct steps, selecting and applying suitable tools to perform specific image manipulations and generate visual rationales $(VR)$. These visual rationales are seamlessly integrated with textual rationales $(TR)$ to construct comprehensive multimodal rationales $(MR)$. By managing the entire process—from action planning to execution—the MLLM ensures consistency and coherence in the reasoning, maintaining a consistent approach across different stages of problem-solving. The overall procedure is illustrated in the left part of Figure \ref{fig:tools}.

\textbf{Action Choices.}
The MLLM designs the chain-of-actions based on the question, similar to how a human would approach a problem by breaking it down into a series of sub-goals $(SG)$. For instance, in the example shown in Figure \ref{fig:tools}, the model might split a question into sub-goals, such as "identify the buses in the image" the model then selects appropriate tools for each sub-goal, such as object detection for identifying the buses. 
To empower the MLLM with capabilities crucial for intricate visual analysis, our Image-of-Thought method incorporates a curated set of actions, each meticulously selected to enhance specific aspects of the image processing task. The implementation details for each $Action$ are provided in Appendix~\ref{appendix:tool}: 
\begin{itemize}
    \item[] \textbf{Segmentation / Edge Detection:} Identifies and delineates clear boundaries within the image, aiding in object recognition and spatial analysis by highlighting edges and contours.
    \item[] \textbf{Geometric Transformations:} Focuses on and magnifies specific areas of interest, facilitating detailed examination of subtle features to improve visual comprehension.
    \item[] \textbf{Dense Object Detection / Referring Object Detection:} Detects objects in densely packed scenes or identifies specific objects based on textual descriptions, linking visual elements accurately with textual queries.
    \item[] \textbf{Spatial Ruler: }Divides the image into quadrants using horizontal and vertical axes, aiding in perceiving spatial relationships and addressing spatial positioning queries accurately.
    \item[] \textbf{Color Space Conversion:} Transforms the image's color space to differentiate objects based on color characteristics, enhancing visual contrast and aiding in effective segmentation and categorization.
\end{itemize}

\textbf{Multimodal Rationales Generation. }
Subsequently, the MLLM generates the commands to execute each $Action$, obtaining the $VR$ for each step. 
Once the $VR$s are obtained from each step, they are integrated with the corresponding $TR$s generated by the MLLM, forming comprehensive $MR$s in the form of the triple <step, visual rationale, textual rationale>. This integration guides the model toward the final decision for the question. Since the visual rationales are generated based on the textual ones, they naturally align.


\subsubsection{Hybrid-Rationale-based Answer Refining}
\label{cpt2-2}
\textbf{Chain-of-Image-Thought.}
After generating the $MR$s, we concatenate them sequentially to create a multimodal rationales series $(MRS)$, which visualizes the internal reasoning process of the MLLM across both aligned modalities. This series provides a transparent way to observe the MLLM's internal logic when addressing each question, making it easier to understand the reasoning behind the model's behavior and determine the accuracy of its answers. 

\textbf{Hybrid Rationale Refining.}
Based on the $MRS$, we feed them back into the MLLM along with the corresponding question. The model then refines its answer based on this series, producing a final output.
Our approach leverages the intrinsic capabilities of MLLMs to refine decision-making processes without additional training or annotated data. This training-free and data-free method uses sophisticated prompting to enable the model to automatically refine its own decisions, enhancing both the accuracy and interpretability of the final output.
Details are illustrated in the right part of Figure~\ref{fig:tools}.

\subsubsection{IoT Algorithm}
As demonstrated in Algorithm~\ref{alg:IoT},  $Q$ denotes the input question, and $I$ represents the input image. The final output is the answer $A$, accompanied by the Multimodal Rationale Series $MRS$, a sequence of multimodal rationales. The question $Q$ is decomposed into a series of sub-goals $SG$. For each sub-goal, an appropriate image processing action $Action$ is selected. The execution of these actions yields Visual Rationales $VR$, while the Multimodal Large Language Model (MLLM) generates the corresponding Textual Rationales $TR$. Each step forms a Multimodal Rationale $MR$, expressed as a triple $\langle SG, VR, TR \rangle$.

\begin{algorithm}[h]
\caption{Image-of-Thought Prompting}
\label{alg:IoT}
\begin{algorithmic}
\State \textbf{Input:} Question $Q$, Image $I$
\State \textbf{Output:} Final Answer $A$, Multimodal Rationale Series $MRS$
\State Initialize $MRS \leftarrow []$ 
\State $\{SG_1, SG_2, \ldots, SG_n\} \leftarrow \text{MLLM.auto\_design}(Q)$ \Comment{Auto-design the IoT processes.}
\For{each sub-goal $SG_i$}
    \State $Action_i \leftarrow \text{MLLM.select\_action}(SG_i)$ 
    \State $VR_i \leftarrow \text{execute}(Action_i, I)$ \Comment{Generate the visual rationale $VR_i$}
    \State $TR_i \leftarrow \text{MLLM.generate\_text\_rationale}(SG_i)$ \Comment{Generate the textual rationale $TR_i$}
    \State $MR_i \leftarrow \langle SG_i, VR_i, TR_i \rangle$ 
    \State $MRS \leftarrow MRS \cup MR_i$ 
\EndFor
\State $MRS \leftarrow \text{MLLM.concat\_rationales}(\{MR_1, MR_2, \ldots, MR_n\})$ 
\State $A \leftarrow \text{MLLM.refine\_answer}(Q, MRS)$ \Comment{Feed the hybrid rationales to generate the final answer $A$}
\State \textbf{Return} $\langle A, MRS \rangle$ 
\end{algorithmic}
\end{algorithm}

\subsection{IoT Prompting Properties}


The Image-of-Thought prompting method offers several properties for facilitating reasoning in multimodality models: 
\begin{enumerate}
    \item Systematic Integration of Image Processing: The IoT method systematically integrates discrete image processing operations within the model's reasoning chains. This integration is critical for grounding the model's reasoning steps in visual reality, ensuring that each decision is substantiated by direct image evidence rather than solely by textual interpretation.
    \item The IoT method equips MLLMs with the tools to dissect visual and textual data elements effectively. This ensures that each question aspect is addressed through a focused analysis, enhancing the model's ability to understand and interpret complex scenes.
    \item The explicit visualization of reasoning steps in the IoT method improves the interpretability of the model's decisions. By directly showing the internal logic of the reasoning process, the method enhances the transparency and understandability of the MLLM's operations, making it easier for human observers to follow and trust the model's conclusions.
\end{enumerate}

\section{Experiments}

\subsection{Experimental Setup}
\label{cpt3:setup}
\textbf{Tasks and Datasets.} We evaluated the IoT prompting on three benchmark datasets from Visual Question-Answering tasks with multiple aspects. 
\textbf{MMBench}\citep{liu2023mmbench} is a systematically designed objective evaluation benchmark to evaluate the different abilities of large vision-language models robustly. It has questions covering 20 different ability dimensions for evaluating vision-language models. Each ability dimension, with the number of questions maintained, is roughly equal. The distribution facilitates a balanced and thorough assessment of these abilities. In line with the initial assessment standards, their metrics determine accuracy by comparing the number of correctly answered questions to the total number of questions.
\textbf{MME}\citep{fu2024mme} is a comprehensive MLLM Evaluation benchmark that measures both the perception and cognition abilities of MLLMs on a total of 14 subtasks. It uses the MME framework to assess model performance via accuracy (Acc), contributing to an overall score across all subtasks.
\textbf{MMVet}\citep{yu2024mm-vet} is an evaluation benchmark that examines MLLMs on complicated multimodal tasks. It defines 6 core Vision-Language capabilities and examines the 16 integrations of interest derived from the capability combination. For evaluation metrics, an LLM-based evaluator for open-ended outputs is proposed. 

\textbf{Baselines. }Given the requirement for conducting image-related operations and reintroducing the image during intermediate stages for multi-step reasoning, we opted to employ a baseline model capable of engaging in multiple conversation rounds. Consider the powerful visual reasoning capability demonstrated by GPT4\cite{gpt4v,gpt4o} and Gemini-Pro\cite{geminiteam2024gemini}, in our experimental setup, we specifically selected the GPT-4 and Gemini-Pro\footnote{Model details: For GPT-4, we select the \textbf{gpt4o} as the baseline model. We choose the \textbf{gemini-pro-vision-1.5} as the baseline model for Gemini-Pro.} models as the baseline models. Subsequently, we integrated the IoT and evaluated the experimental results of these models.

\textbf{Toolbox Selection. }
To facilitate the sequential inference process involving image manipulation to extract visual rationales, we use an extendable toolbox including two image processing models \textbf{FastSAM}\citep{zhao2023fast} and \textbf{GroundingDINO}\citep{liu2023grounding} respectively that possess the capability to encode textual information with textual rationales, and a \textbf{Python Imaging Library}\citep{clark2015pillow} to extract the primary visual rationale, we listed the details of the tools in Appendix~\ref{appendix:tool}.

\textbf{Implementation. }
We establish specific prompts that guide the model's thought process across multiple stages to enable step-by-step reasoning. To extract the necessary steps and target objects for image processing, the prompt allows the model to independently determine the sequence and order based on the input image. For instance, in the case of the MMBench dataset, the prompt is formulated as \textit{"Give the images, questions, and options, please think step-by-step about the image features to answer the question. The MLLMs have access to various image processing operations that can assist the models in addressing the query."} More details are presented in the Appendix~\ref{appendix:prompt}.
To ensure consistent and reliable results, we utilize GPT-4's API with a temperature setting of 0 and a maximum token limit of 500 to promote stability in the generated outputs.

\subsection{Results}
The results of Image-of-Thought prompting on the MMBench and MME benchmarks are listed in Table~\ref{result-table_0}. Due to the limited size of input images for Gemini-Pro, it is a little difficult to input a vision rationale series during experiments on MMVet (which has larger images compared to MMBench and MME), so we only experiment the GPT-4o with IoT on MMVet. The result on MMVet of GPT-4o is listed in Table~\ref{result-table_3}. The MMBench evaluation method is the accuracy of the exact match answer. MME evaluation encompasses aggregating the accuracy values across all categories. And the MMVet bench employs GPT-4 as an evaluator to determine the correctness score of the predicted answer.

\begin{figure}
    \centering
    \includegraphics[width=0.8\textwidth]{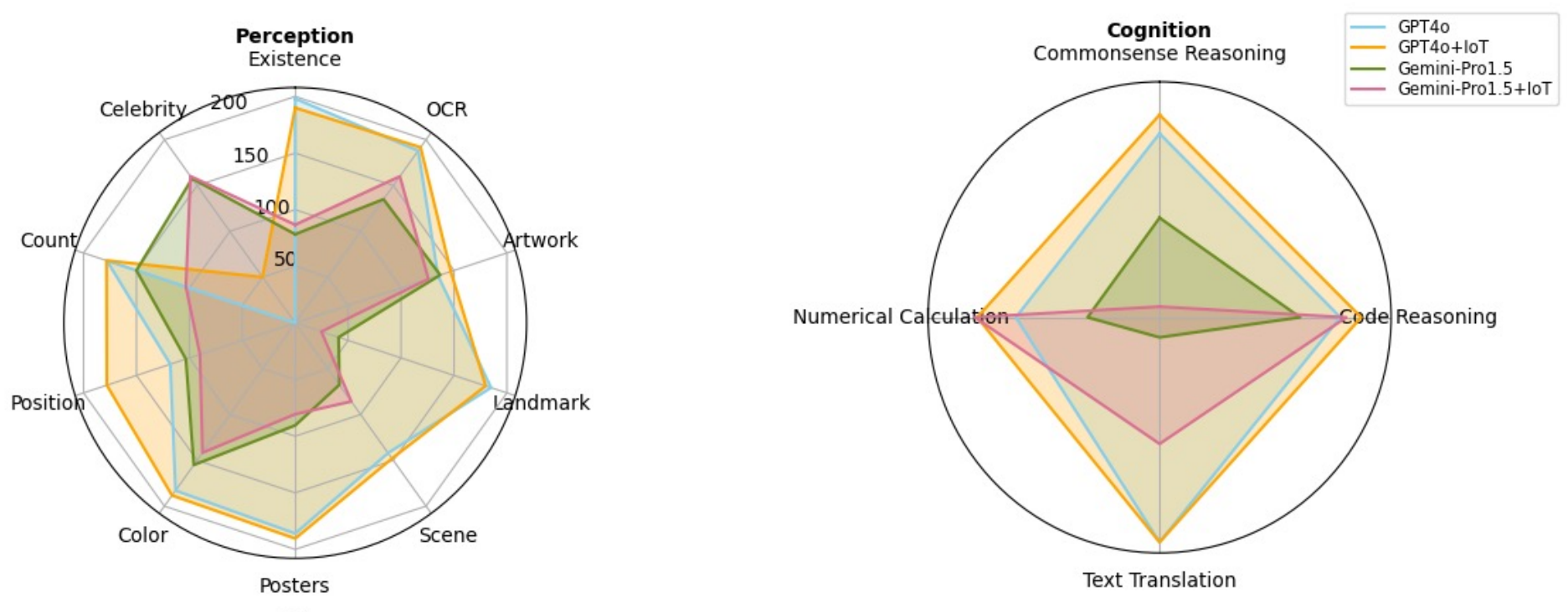}
    \caption{Accuracy values for different categories of the MME benchmark. IoT has enhanced the performance of both GPT-4o and Gemini-Pro-1.5, particularly demonstrating a greater impact in the reasoning and position categories.}
    \label{fig:mme_c}
\end{figure}

While IoT exhibits improvements across the overall evaluation criteria, demonstrating a more considerable reasoning category enhancement. Figure \ref{fig:mme_c} illustrates the results of various categories in the MME benchmark. We can find that the IoT significant improved more the model's performance on Cognition task (GPT-4o: +5.6\%, Gemini-pro-1.5: +23.9\%, more details in Appendix~\ref{appendix:detail}), which imply that the utilization of IoT potentially provides the model with additional contextual information and enhances its ability to process and reason with complex cognitive tasks more effectively. We hypothesize that this is attributed to IoT's explicit expansion of the model's question-answering process, thereby enhancing the model's reasoning capabilities.

    

\begin{table}
  \caption{Image-of-Thought Prompting Performance on MMBench(dev) and MME (The asterisk (*) denotes that the corresponding answer was obtained through our independent reproduction.)}
  \label{result-table_0}
  \centering
  \begin{tabular}{lccccc}
    \toprule
    
    \textbf{Model}  & \textbf{Params}  & \textbf{ MME Total }  & \textbf{MMBench(dev)}   \\
    \midrule
    MiniGPT-4\cite{zhu2023minigpt}    &  8B & 725.95 & 24.3 \\
    LLaVA-v1.5-13B\cite{liu2023improved}     &  13.4B & 1615.6 & 69.2  \\
    InternLM-XComposer2-VL\cite{zhang2023internlm} & 7B &2242.71   & 79.5  \\
    Qwen-VL-Plus\cite{bai2023qwen}   & - & 2183.39 & 66.2 \\
    Qwen-VL-Max   &  - & 2433.61 & 78.1   \\     
    \midrule
    Gemini-Pro-1.5*     &  -    & 1571.4 & 71.0  \\
    Gemini-Pro-1.5+IoT*     &  -    & 1686.1 & 79.6  \\
    GPT4o*     &  -    & 2254.1 & 86.2  \\
    GPT4o + IoT*  &  -   & \textbf{2396.1} & \textbf{87.6} \\
    \bottomrule
  \end{tabular}
\end{table}

    

\begin{figure}
    \centering
    \includegraphics[width=0.8\textwidth]{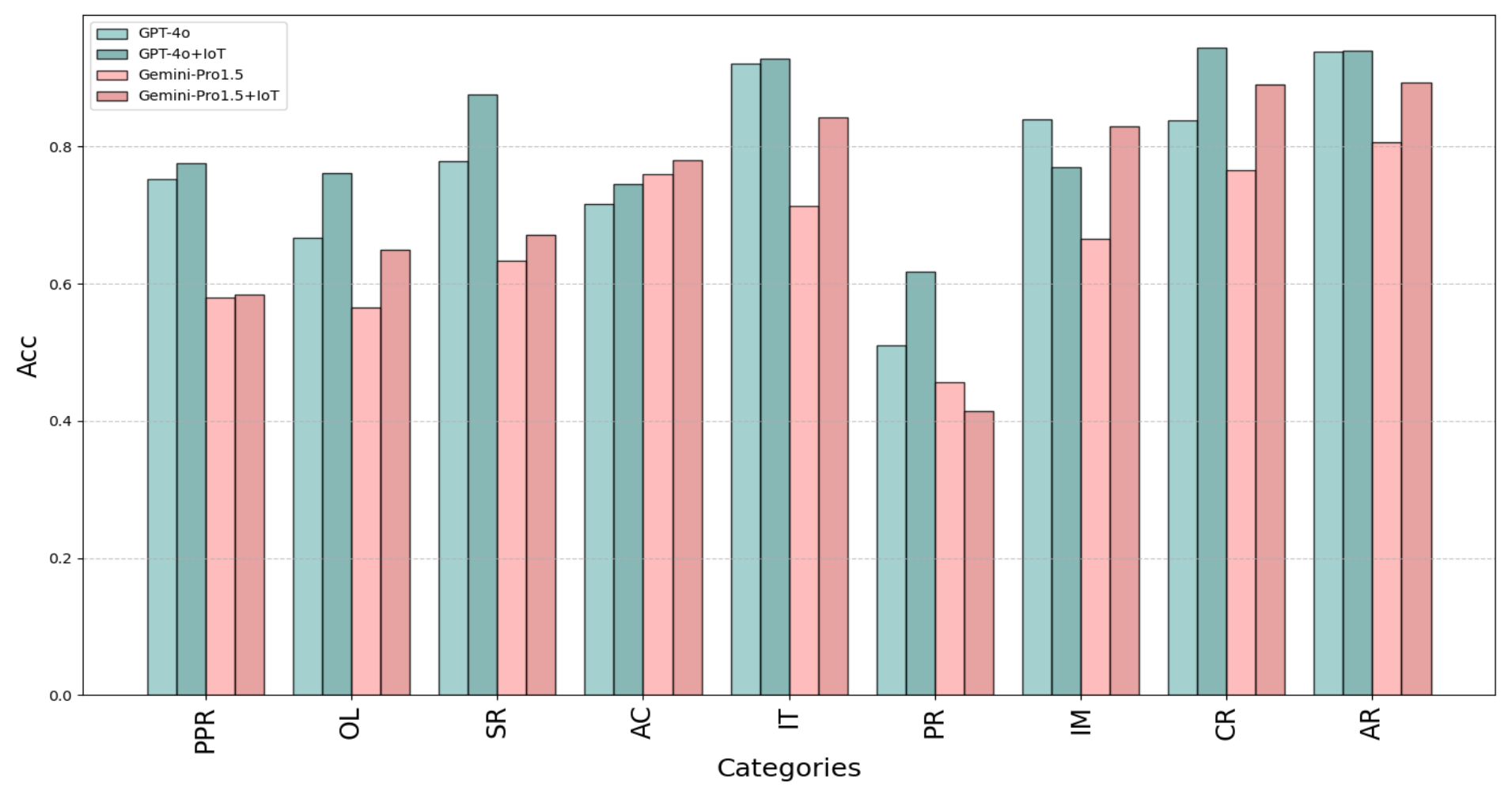}
    \caption{Accuracy values for some categories in the MMBench Dev. PPR: Physical Property Reasoning. OL:Object Localization. SR: Spatial Relationship. AC: Attribute Comparison. IT: Image Topic. PR: Physical Relation. CR: celebrity Recognition. AR: Attribute Recognition. IoT leads to improved performance across most categories, particular in Object Location, Spatial Relation, Attribute Comparison and Celebrity Recognition.}
    \label{fig:mmb}
\end{figure}

The results for some categories in the MMBench dataset are presented in Figure \ref{fig:mmb}. For GPT-4o, the utilization of IoT leads to improved performance across lots of categories, particularly when the model needs to compare different objects through reasoning. Compared to GPT-4o, the model demonstrates significant performance gains in the Physical Relation category by 20.98\%, in the Object Location category by 16.34\%, and in the Spatial Relation category by 12.45\%. When comparing the performance of IoT with Gemini, it is evident that IoT leads to improved performance across more categories, especially, there is a significant increase of 30.24\% in Action Recognition, 24.81\% in Image Emotion and 17.92\% in Image Topic. This enhancement in perception and reasoning capabilities is likely attributed to the model's utilization of actions, which emphasize the target objects within the image. Detailed results and specific numerical values for other categories are in Appendix~\ref{appendix:detail}.

However, there is a certain loss in performance for some certain categories for both models. We speculate that this could be due to the multi-image processing approach of GPT-4o and Gemini-Pro-1.5, which might result in reduced resolution and information about the original image and vision rationales. We have provided examples of such errors in the Appendix~\ref{appendix:case}, along with further analysis.

\begin{wraptable}{r}{8cm}
  \caption{Ablation Study on MMVet}
  \label{result-table_3}
  \centering
  \begin{tabular}{l@{}c@{}c@{}c}
    \toprule
    
    \textbf{Category}  & \textbf{GPT-4o}  & \textbf{ +text rationale }  & \textbf{+IoT}   \\
    \midrule
    Recognition  &64.7   & 65.0 & \textbf{65.6}  \\
    OCR &80.1 &82.9 &\textbf{83.3} \\
    Konwledge &57.0 &56.2 &\textbf{60.0} \\
    Language Generation &\textbf{61.5} &60.9 &61.4 \\
    Spatial Awareness &72.0  &74.3  &\textbf{77.9}  \\
    Math  &85.4   &\textbf{92.3}  &91.9 \\
    \midrule
    \textbf{Total} &70.5 &70.9 &\textbf{72.2} \\

    \bottomrule
  \end{tabular}
\end{wraptable} 

\subsection{Rationale Ablation Study}
In this section, we analyze the impact of hybrid rationales.
As depicted in Table~\ref{result-table_3}, our observations indicate that removing visual rationales but only keeping the text rationales in a decline in the model's performance on the MMVet dataset, particularly in the Knowledge and Spatial Awareness categories(Knowledge: -6.76\%, Spatial Awareness: -4.85\%). This suggests that textual rationales alone may not effectively capture image-specific knowledge and spatial information. In other words, the model with IoT usually works better on spatial tasks, which can also be found in the MMBench spatial relation task. 

Interestingly, in the case of Math questions, the performance of pure textual rationales surpasses that of hybrid rationales and zero-shot approaches. This intriguing finding suggests that the visual rationales may introduce additional symbols, potentially leading to visual hallucination and consequently impacting the model's ability to reason effectively in mathematical contexts.

    



\subsection{Action Statistics}
\begin{figure}
    \centering
    \includegraphics[width=0.8\textwidth]{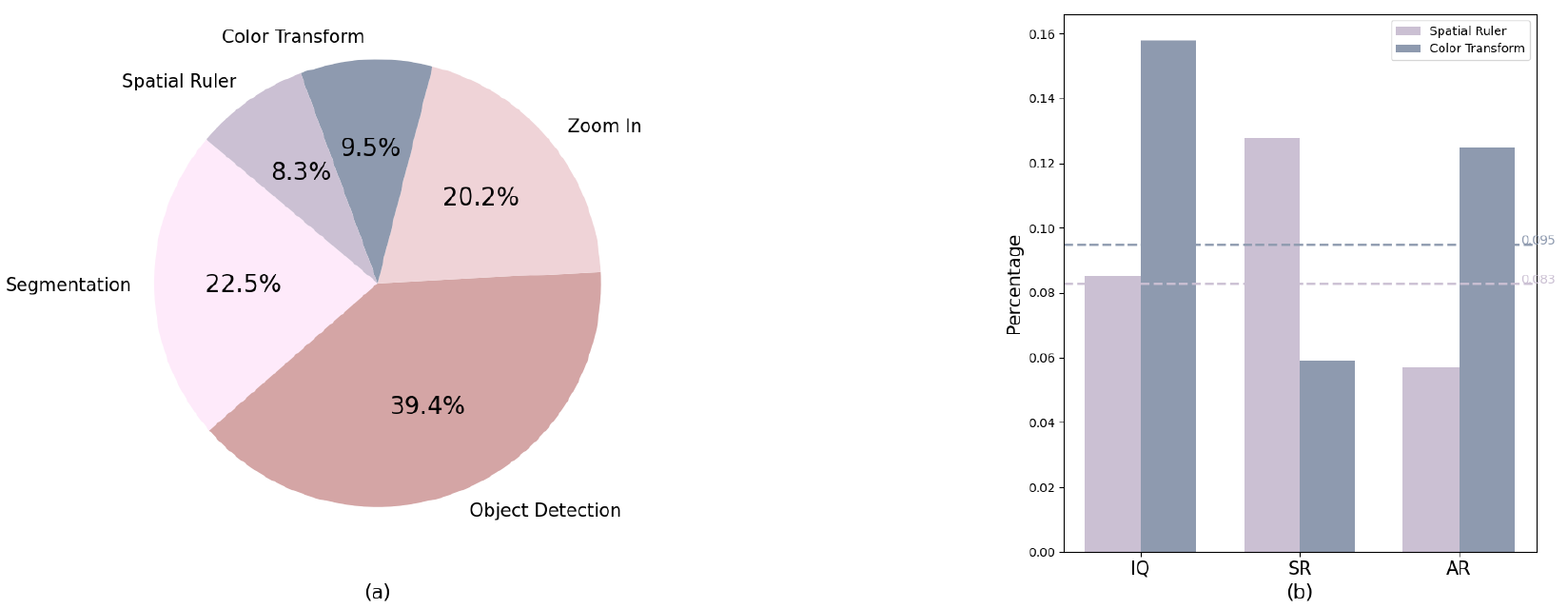}
    \caption{(a): Action counts on MMBench Dev benchmark. The most common use action is the object detection and followed by segmentation and zoom in. (b): IQ: Image Quality. SR: Spatial Relation. AR: Attribute Recognition. These categories utilize more Color Transform and Spatial Ruler actions.}
    \label{fig:action}
\end{figure}

We primarily analyze the actions employed within these datasets in the MMBench Dev datasets, which contain over four thousand data points. In MMBench, the models' most commonly employed action is object detection, followed by segmentation and zoom-in, as depicted in Figure \ref{fig:action}(a). This indicates a resemblance between the model's approach to accomplishing the VQA task and human thought processes. Specifically, the model prioritizes locating the target object (object detection action) before attempting to separate it from the background, which may achieved by segmentation or zoom-in. As for the other two actions, spatial ruler and color transform, which operate on the entire scene rather than individual objects, the model tends to employ them less frequently. Additionally, Color Transform may be used to drop the color features in the image if the question does not require it. While the Spatial Ruler provides the model with a helpful guideline for determining the positional relationships of objects, the Object Detection action, to some extent, aids the model in enhancing spatial perception when localizing multiple target objects.

However, specific categories, such as image-quality and spatial-relation questions, benefit from using Color Transform and Spatial Ruler actions, as indicated by the action utilization percentages displayed in Figure \ref{fig:action}(b). In the case of Image Quality questions, Color Transform proves valuable by enabling a focused examination of factors like brightness or noise, providing insights for addressing such questions effectively. Moreover, Spatial Relation tasks typically involve comprehending the arrangement, position, and relationships among objects or elements within an image. This aligns with human thinking patterns, as individuals often employ measurements and qualifications to obtain more precise outcomes in spatial analysis.
More details of action statistics can be found in Appendix~\ref{appendix:action_count}.

\section{Conclusion}
This paper introduced the Image-of-Thought (IoT) prompting method, a novel train-free approach designed to enhance multimodal large language models (MLLMs) for visual question-answering tasks. Our method seamlessly integrates discrete image processing actions within the reasoning chains generated by the language models, thereby improving both accuracy and interpretability in multimodal reasoning. By considering textual and visual rationales, IoT prompting provides a more holistic approach to understanding and responding to visual inputs.
Across various benchmarks, IoT prompting reduces hallucinations and reliance on textual information in zero-shot learning scenarios to some extent.

\section{Limitations}
Despite its advantages, the IoT method has several limitations that must be acknowledged. While IoT emulates the human multimodal reasoning process, its reliance on hybrid rationales does not equate to genuine cognitive reasoning by the model. IoT's step-by-step intelligence is restricted to specific process areas and does not encompass the full spectrum of cognitive reasoning capabilities. 
However, with the development of MLLMs
The flexibility inherent in the IoT method allows for selecting appropriate tools at each process step, enhancing adaptability across different multimodal scenarios. For instance, in future works, the IoT method in robotics research can autonomously design action sequences using plausible hybrid rationales.

\bibliographystyle{plainnat}
\bibliography{neurips_2024}

\newpage
\appendix
\definecolor{celadon}{rgb}{0.67, 0.88, 0.69}
\definecolor{babypink}{rgb}{0.96, 0.76, 0.76}

\section*{Appendix}
In this section, we present additional implementation details, experiment results and cases.

\section{Implemented Toolbox Details}
\label{appendix:tool}

We listed the details of the toolbox used for extract visual rationales in the paper:
\begin{enumerate}
    \item \textbf{FastSAM}\citep{zhao2023fast} is a CNN Segment Anything Model trained using only 2\% of the SA-1B dataset published by SAM authors. FastSAM achieves comparable performance with the SAM method\citep{kirillov2023segment} at 50× higher run-time speed. As a result, we employ FastSAM as our preferred segmentation tool. To obtain the necessary text prompt for segmentation, we extract relevant information from the text rationale.
    \item \textbf{GroundingDINO}\citep{liu2023grounding} is an open-set object detector which can detect arbitrary objects with human inputs such as category names or referring expressions. When the action is identified as object detection or zooming in, the first step is to extract the target object from the text rationale. Subsequently, by inputting the target object into the model, we obtain the coordinates of the bounding box. For object detection purposes, we draw the bounding box around the identified object. In the case of a zoom-in action, we zoom in on the specified region based on the provided coordinates.
    \item \textbf{Python Imaging Library}\citep{clark2015pillow} provide a solid foundation for a general image processing tool. In order to perform color transformation and incorporate a spatial ruler into an image, we have developed functions based on this Python library. These functions enable the conversion of the image to grayscale, or the addition of a spatial ruler onto the image.
\end{enumerate}

\section{Additional Result and Analysis}
\subsection{Detailed Results for MMBench(dev), MME and MMVet}
\label{appendix:detail}

\textbf{MMbench(dev)}
Tabel~\ref{appendix_1} shows all categories' accuracy of MMBench(dev). We observe that for GPT-4o, the categories marked in green demonstrate significant improvement when coupled with IoT. Conversely, the categories labeled in red show lower performance compared to the baseline. In contrast, for Gemini-pro-1.5, the majority of these categories display an increase in performance. This discrepancy may be attributed to the inherent capabilities of the models themselves and their respective approaches to processing images and text.

\begin{table}[h]
  \caption{All categories' accuracy in MMBench(dev)}
  \label{appendix_1}
  \centering
  \begin{tabular}{lcccc}
    \toprule
    
    \textbf{Category}  & \textbf{GPT-4o}  & \textbf{Gemini-Pro} & \textbf{GPT-4o+IoT} & \textbf{Gemini-Pro+IoT}  \\
    \midrule
    identity reasoning     &  0.994 &  0.830 & 0.966 & 0.943     \\
    physical property reasoning     &  0.753 &  0.580 & 0.776& 0.584      \\
    function reasoning     &  0.927 &  0.789 & 0.885 & 0.862    \\
    \rowcolor{yellow!20}object localization     &  0.667 &  0.565 &  \cellcolor{celadon} 0.762 & 0.650     \\
    structuralized imagetext understanding     &  0.907 &  0.606 & 0.897 & 0.776    \\
    attribute recognition     &  0.939 &  0.807 & 0.940 & 0.894     \\
    future prediction     &  0.738 &  0.515 & 0.738& 0.561     \\
    \rowcolor{yellow!20}spatial relationship     &  0.779 &  0.633 & \cellcolor{celadon} 0.876 & 0.672    \\
    image scene     &  0.990  &  0.803 & 0.980 & 0.911   \\
    \rowcolor{yellow!20}image quality     &  0.547 &  0.567 & \cellcolor{celadon}0.700 & 0.380     \\
    action recognition     &  0.967  &  0.721 & 0.967 & 0.939    \\
    attribute comparison     &  0.716&  0.760 & 0.745 & 0.780     \\
    image topic     &  0.921  &  0.714 & 0.929 & 0.842    \\
    nature relation     &  0.983&  0.866 & 0.978 & 0.911     \\
    \rowcolor{yellow!20}physical relation     &  0.510  &  0.457 & \cellcolor{celadon}0.617 & 0.415    \\
    \rowcolor{yellow!20}social relation     &  0.924&  0.750 & \cellcolor{babypink}0.860 & 0.895     \\
    celebrity recognition     &  0.838 &  0.765 & 0.945 & 0.891   \\
    image style     &  0.943  &  0.708 & 0.877 & 0.807    \\
    ocr     &  0.987 & 0.846  & 0.962  &  0.872  \\
    \rowcolor{yellow!20}image emotion     &  0.840  &  0.665 & \cellcolor{babypink}0.770 & 0.830    \\

    \bottomrule
  \end{tabular}
\end{table}

\textbf{MME}
Tabel~\ref{appendix_2} shows all categories' accuracy of MME. IoT has had a substantial positive impact on the performance of cognitive tasks for both GPT-4o (+5.6\%) and Gemini-pro-1.5 (+23.9\%). These findings suggest that incorporating IoT technology potentially offers the model access to supplementary contextual information, thereby augmenting its capacity to effectively process and reason with intricate cognitive tasks.

\begin{table}[h]
  \caption{All categories' accuracy in MME}
  \label{appendix_2}
  \centering
  \begin{tabular}{lllll}
    \toprule
    
    \textbf{Category}  & \textbf{GPT-4o}  & \textbf{Gemini-Pro} & \textbf{GPT-4o+IoT}  & \textbf{Gemini-Pro+IoT} \\
    \midrule
    \multicolumn{5}{l}{\textbf{Perception}}         \\
    \cmidrule(r){1-5}
    existence     &  200.0 &78.3  &191.2  &86.7 \\
    count & 180.0  & 150.3 & 179.3 &165.1\\
    \rowcolor{yellow!20}position & 118.3 &103.3  &\cellcolor{celadon}178.3 &90.0\\
    color &185.0  &155.0 &188.3 &141.7\\
    posters & 189.5 &90.5 & 190.5 &80.6\\
    celebrity &0.0 &157.6 & 50.1 &160.2\\
    scene & 147.2 &67.5 &148.0 &85.8\\
    landmark & 189.3  &41.0 &180.0 &24.8\\
    artwork & 141.1 &137.0 & 146.9 &125.8\\
    OCR & 192.5 &135.0 & 191.7 &160.0\\
    \midrule
    \multicolumn{5}{l}{\textbf{Cognition}}                   \\
    \cmidrule(r){1-5}
    \rowcolor{yellow!20}commonsense reasoning & 178.5  &123.6 &\cellcolor{celadon}186.4 &69.3\\
    \rowcolor{yellow!20}numerical calculation  & 155.0  & 107.5 &\cellcolor{celadon}176.8 &177.5\\
    text translation & 200.0 &75.0 &200.0 &140.0\\
    \rowcolor{yellow!20}code reasoning & 177.5 & 150.0 &\cellcolor{celadon}188.0 &178.7\\

    \bottomrule
  \end{tabular}
\end{table}

\textbf{MMVet}
Tabel~\ref{appendix_3} shows all categories' results of MMVet which are evaluated by GPT4. The results obtained indicate that the integration of IoT can also enhance the model's capabilities in OCR, Knowledge, Spatial Awareness, and Math. Particularly, the notable improvement in the performance of the model on Spatial Awareness problems(+8.19\%) and Math problems(+7.61\%) suggests that the IoT will be particularly effective on some reasoning tasks.

\begin{table}[h]
  \caption{All categories' results of MMVet}
  \label{appendix_3}
  \centering
  \begin{tabular}{lll}
    \toprule
    
    \textbf{Category}  & \textbf{GPT-4o}  & \textbf{GPT-4o+IoT}  \\
    \midrule
    Recognition     &  64.7   &65.6  \\
    OCR     &  80.1   &\cellcolor{celadon}83.3  \\
    Knowledge     &  57.0  & \cellcolor{celadon}60.0 \\
    Language Generation     &  61.5 & 61.4   \\
    Spatial Awareness  &  72.0 & \cellcolor{celadon}77.9 \\
    Math     &  85.4 &\cellcolor{celadon}91.9  \\

    \bottomrule
  \end{tabular}
\end{table}

\subsection{Overall tool selection statistic}
\label{appendix:action_count}
\begin{figure}[h]
    \centering
    \includegraphics[width=1.0\linewidth]{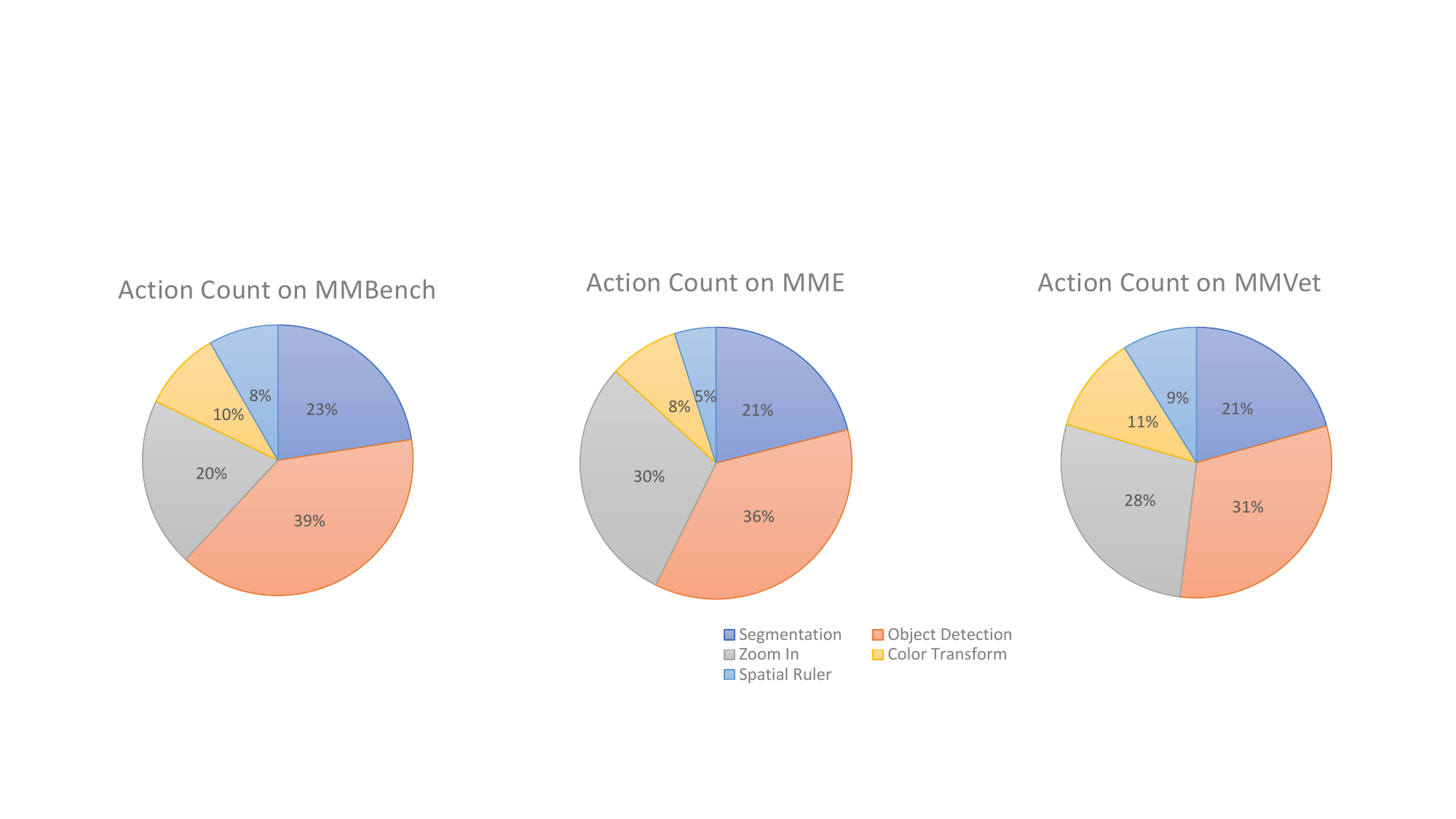}
    \caption{Action Counts in three benchmark}
    \label{fig:action_c}
\end{figure}

Figure~\ref{fig:action_c} shows the counts percentage of the five actions used in each benchmark. Across these three datasets, the most frequently utilized action is Object Detection, followed by Segmentation and Zoom In. Spatial Ruler and Color Transform, on the other hand, are the least employed actions.

\section{Case Studies}
\label{appendix:case}
\begin{figure}[!ht]
    \centering
    \includegraphics[width=1.0\linewidth]{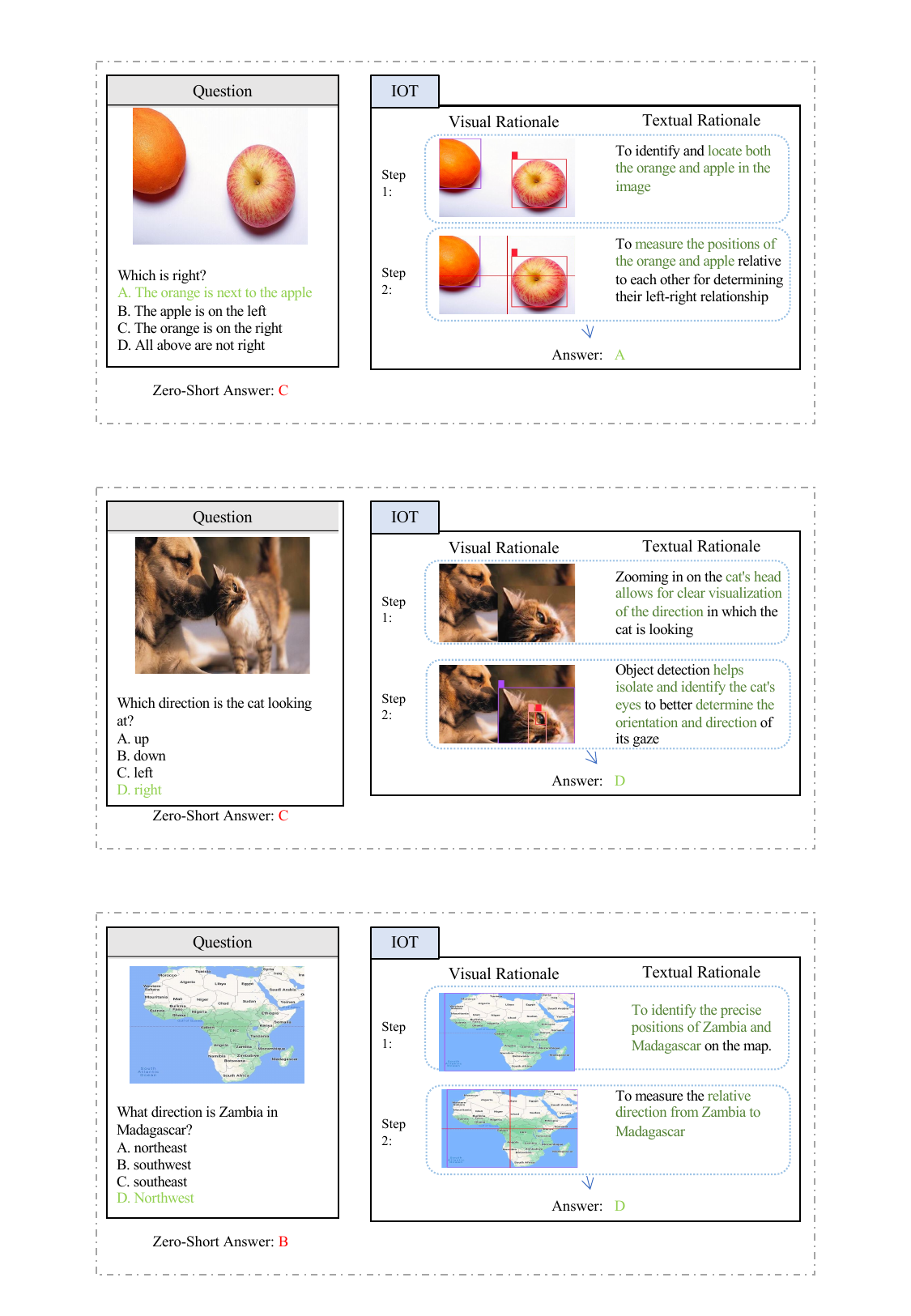}
    \caption{Cases that GPT-4o-IoT are correct, but GPT-4o zero-short are incorrect.}
    \label{fig:case1}
\end{figure}

\begin{figure}[ht]
    \centering
    \includegraphics[width=0.9\linewidth]{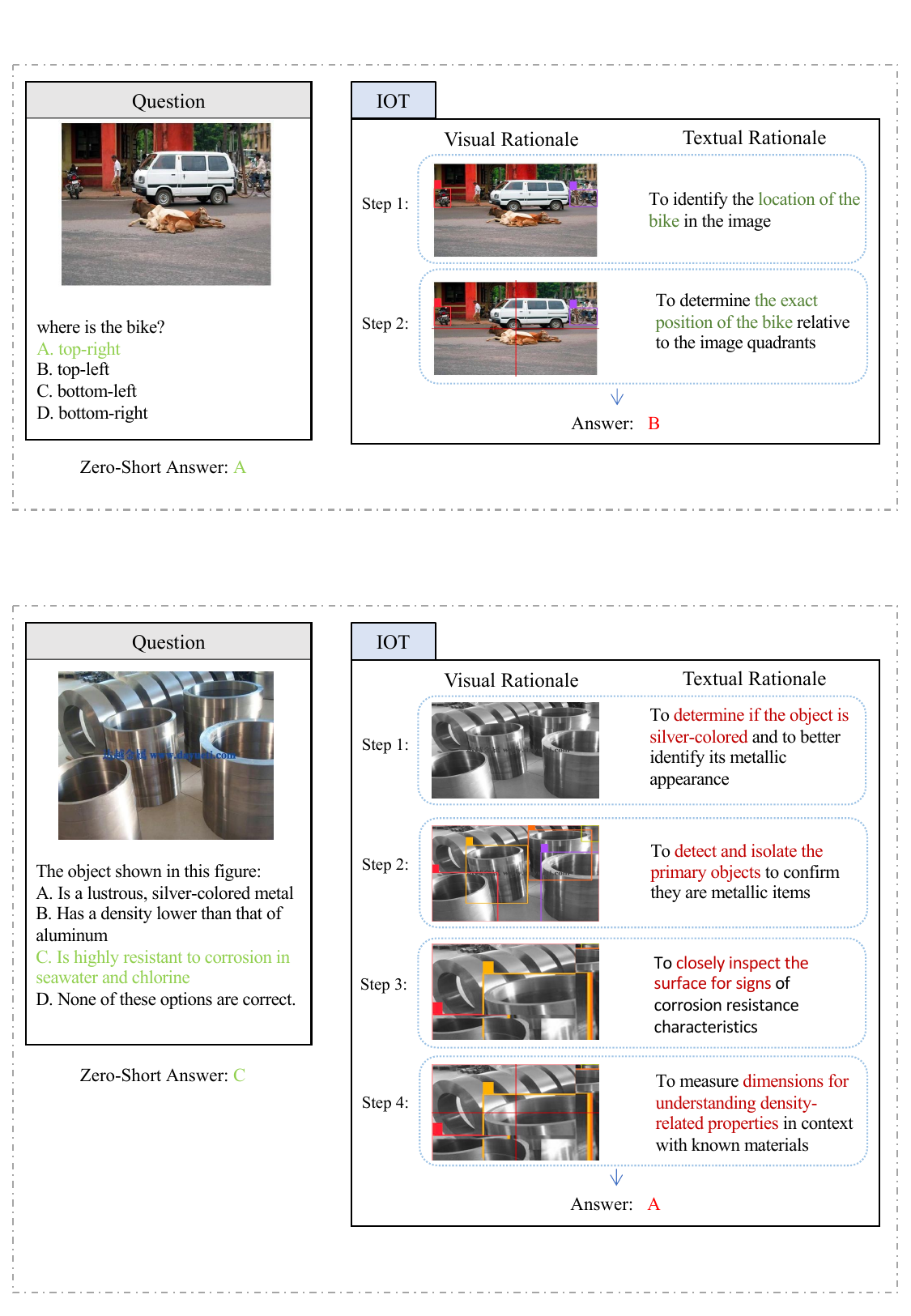}
    \caption{Cases that GPT-4o-IoT are incorrect, but GPT-4o zero-short are correct.}
    \label{fig:case2}
\end{figure}

\begin{figure}[ht]
    \centering
    \includegraphics[width=0.9\linewidth]{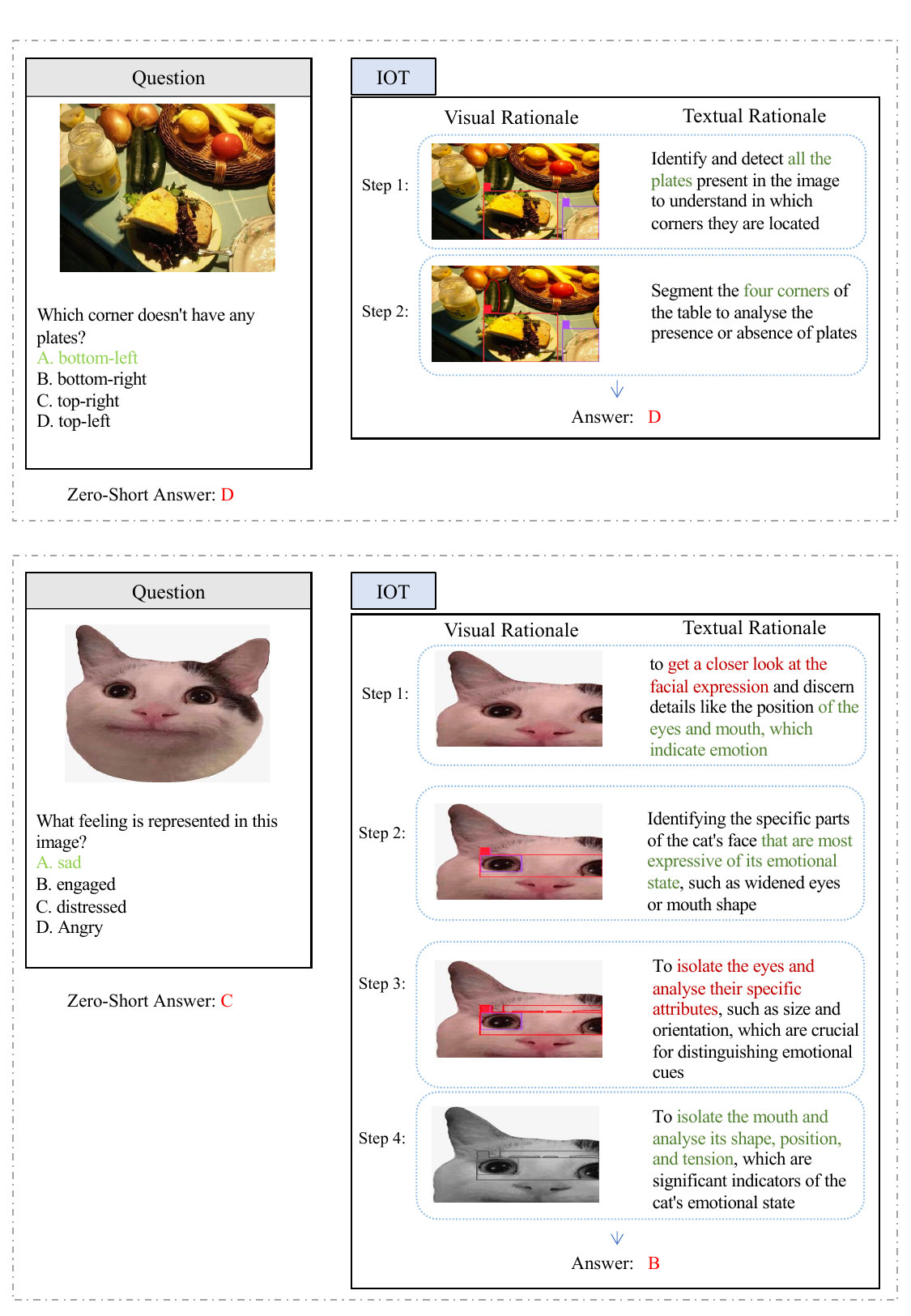}
    \caption{Cases that both GPT-4o-IoT and GPT-4o zero-short are incorrect.}
    \label{fig:case3}
\end{figure}

In order to better explore the effectiveness of the IoT prompting method, we randomly selected three situations of cases in the MMB-dev dataset. 
Specifically, we randomly choose "IoT:\faCheck, Zero-shot:\faTimes", "IoT:\faTimes, Zero-shot:\faCheck"  and "IoT:\faTimes, Zero-shot:\faTimes" three types of cases with their corresponding answers.
Figure~\ref{fig:case1} illustrates the cases in which IoT gets the correct answer with reasonable and clear hybrid rationales.
Figure~\ref{fig:case2} demonstrates that the IoT method may select the wrong answer with visual or textual rationales hallucination.
In addition, Figure~\ref{fig:case3} shows that IoT and Zero-shot methods cannot choose the correct answer for complex scenarios or internet meme images.

\begin{figure}[ht]
    \centering
    \includegraphics[width=1.0\linewidth]{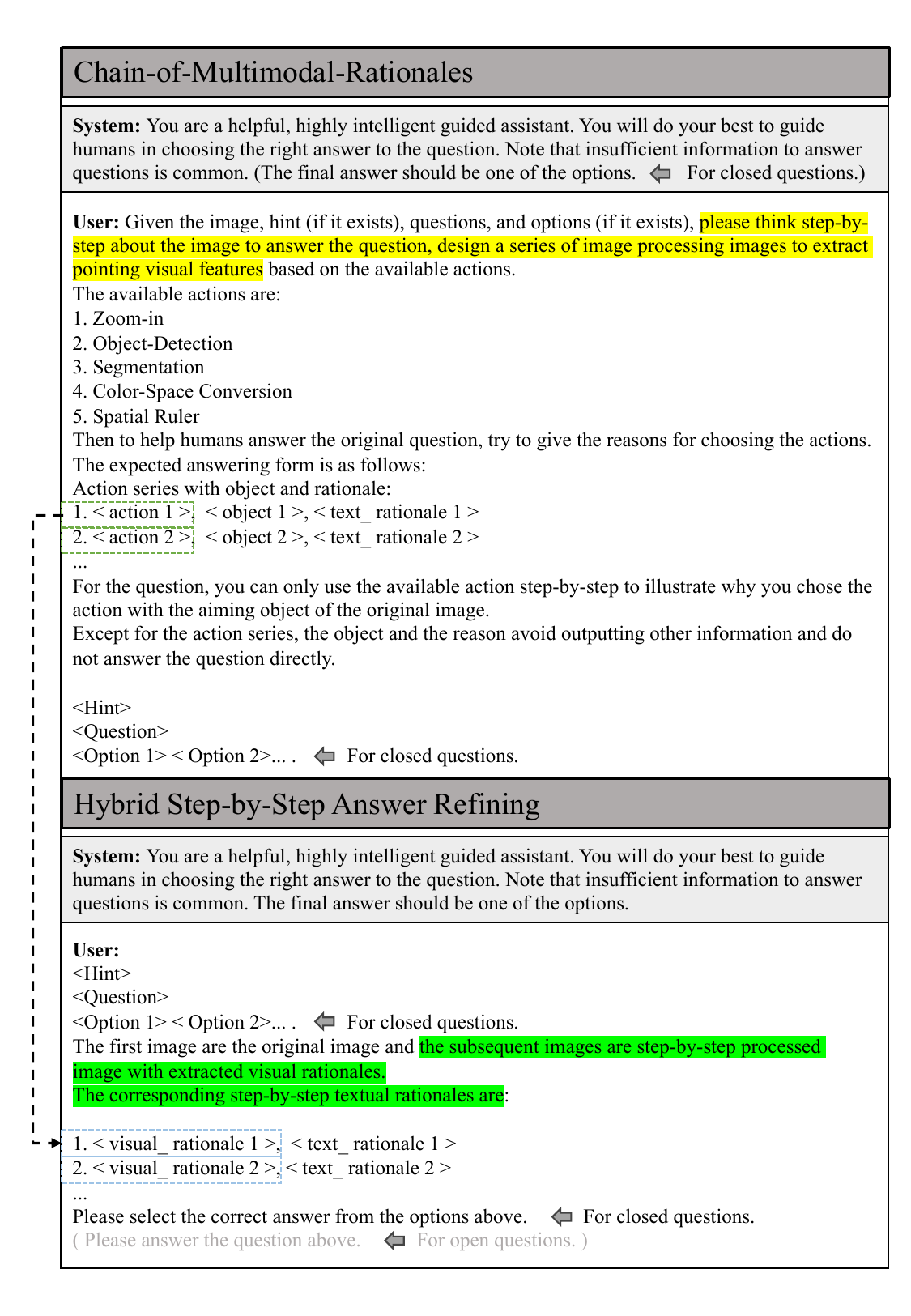}
    \caption{The complete prompts of our Image-of-Thought prompting. The Upper part corresponds to Section 2.1 for a chain of multimodal rationale generation, and the bottom part expresses Section 2.2 for Hybrid step-by-step rationale answer refining. The yellow background highlights the chain-of-action generation prompt, and the green background illustrates the refinement of the hybrid rationale prompt.}
    \label{fig:prompt}
\end{figure}

\section{Detailed Prompt}
\label{appendix:prompt}
This section introduces the details of the proposed IoT prompting.
We employ GPT-4o\citep{gpt4o} and Gemini-pro-1.5\citep{geminiteam2024gemini} to generate the actions and textual rationales. Then, the available tools extract the visual rationale. Finally, GPT-4o and Gemini-pro-1.5 generate the required answer using hybrid rationales.
Figure \ref{fig:prompt} illustrates complete prompts of IoT in the paper.



\end{document}